\documentclass{article}

\usepackage{microtype}
\usepackage{graphicx}
\usepackage{booktabs} 

\usepackage{hyperref}

\usepackage{balance}
\usepackage{stfloats}
\usepackage{subfig}
\usepackage{multirow}
\usepackage{multicol}
\usepackage{makecell}
\usepackage{etoolbox}
\usepackage{enumitem}
\usepackage{xspace}
\usepackage{algorithm}
\usepackage{amsmath}
\usepackage[table]{xcolor}
\usepackage{array, xcolor, pifont}
\usepackage{tcolorbox,tikz}
\usepackage[accepted]{mlsys2025}

\definecolor{aliceblue}{rgb}{0.94, 0.97, 1.0}
\tcbset{width=1\columnwidth,boxrule=1pt,colback=aliceblue,arc=1pt,left=2pt,right=2pt,boxsep=0.5pt}

\newcommand{\ignore}[1]{}
\newcommand{\ours}[0]{HELIOS\xspace}
\newcommand{\review}[1]{\color{red}#1 \color{black}}

\usepackage{tikz}
\newcommand*\bcircled[1]{\tikz[baseline=(char.base)]{
            \node[shape=circle,draw,inner sep=1pt,fill=black, text=white] (char) {#1};}}

\definecolor{OliveGreen}{HTML}{135D66}
\definecolor{mycolor}{HTML}{EEEEEE}

\usepackage{tcolorbox}
\newtcolorbox{hintbox}[2][]
{
  colframe = OliveGreen!100,
  colback  = OliveGreen!5,
  boxsep=2pt,
  boxrule=0.1mm,
  titlerule=0mm,
  width=\dimexpr\columnwidth\relax, 
  coltitle = blue!20!black,
  title    = #2,
  #1,
}

\newcommand{\greentick}{{\color{teal}\large\ding{51}}}

\mlsystitlerunning{\ours{}~: Adaptive Model And Early-Exit Selection for Efficient LLM Inference Serving}

\begin{document}

\twocolumn[
\mlsystitle{\ours{}~: Adaptive Model And Early-Exit Selection\\for Efficient LLM Inference Serving}



\mlsyssetsymbol{equal}{*}

\begin{mlsysauthorlist}
\mlsysauthor{Avinash Kumar}{ut}
\mlsysauthor{Shashank Nag}{ut}
\mlsysauthor{Jason Clemons}{nv}
\mlsysauthor{Lizy John}{ut}
\mlsysauthor{Poulami Das}{ut}
\end{mlsysauthorlist}

\mlsysaffiliation{ut}{Department of Electrical and Computer Engineering, The University of Texas at Austin}
\mlsysaffiliation{nv}{NVIDIA}

\mlsyscorrespondingauthor{Avinash Kumar}{avinkumar@utexas.edu}
\mlsyscorrespondingauthor{Shashank Nag}{shashanknag@utexas.edu}

\mlsyskeywords{Machine Learning, MLSys}

\vspace{0.1in}
\begin{abstract} 
Early-Exit Large Language Models (EE-LLMs) enable high throughput inference by allowing tokens to exit early at intermediate layers. However, their throughput is limited by the computational and memory savings. Existing EE-LLM frameworks rely on a single model and therefore, their token generation latencies are bottlenecked by tokens that do not exit early and traverse additional layers. Moreover, early exits are only known at runtime and depend on the request. Therefore, these frameworks load the weights of all model layers even though large portions remain unused when tokens exit early. The lack of memory savings limit us from scaling the batch sizes. 

We propose {\em \ours}, a framework that
improves both token generation latency and batch sizes to
enable high-throughput in EE-LLMs. \ours{} exploits two insights. \textit{First}, early exits are often complimentary across models, tokens that do not exit early on one model often take an early-exit on another. \ours{} employs multiple models and dynamically switches between them to collectively maximize the number of tokens that exit early, and minimize token generation latencies. \textit{Second}, even when a predicted token does not exit early due to poor confidence, it often remains unchanged even after additional layer traversal. \ours{} greedily allows such tokens to exit early and only loads the weights of the most likely to be used layers, yielding memory savings which is then re-purposed to increase batch sizes. \ours{} employs real-time profiling to accurately identify the early-exit distributions, and adaptively switches between models by tracking tokens in real-time to minimize the performance degradation caused by greedy model loading and exiting. Our evaluations show that~\ours{} achieves $1.48\times$ higher throughput and $15.14\times$ larger batch size compared to existing EE-LLM frameworks.

\end{abstract}
]

\printAffiliationsAndNotice{}

\section{Introduction}

Large Language Models (LLMs) presents critical \textit{throughput} concerns, particularly as we adopt larger and complex models to produce more accurate and nuanced responses. Inference throughput can be increased by reducing token generation latencies and increasing batch sizes or number of concurrent requests served. 
{\em Early-Exit LLMs (EE-LLMs)} are a class of LLMs that allow tokens to exit at specific intermediate layers if their probability meets a confidence threshold. By skipping layers for simple tokens, EE-LLMs reduce the latency of token generation and improves throughput.


\begin{figure}[!t]
  \centering
  \includegraphics[width=1.0\columnwidth]{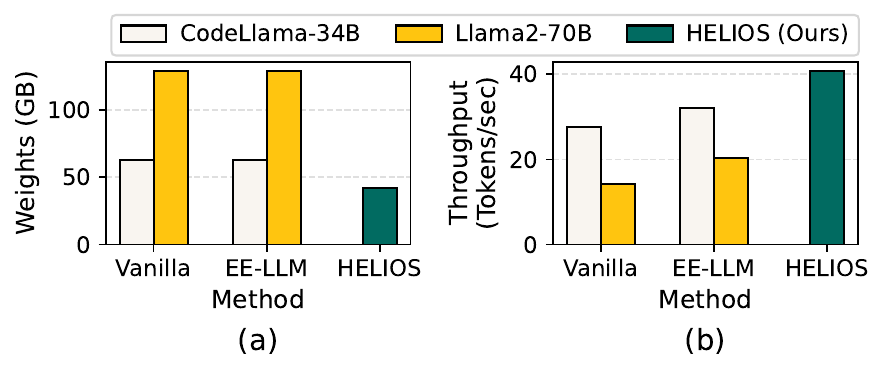}
  \vspace{-0.30in}
   \caption{(a) Memory required to store model weights and (b) throughput of CodeLlama-34B and Llama2-70B models for vanilla auto-regressive decoding, EE-LLM, and~\ours{} on ShareGPT~\cite{sharegpt}. By using multiple LLMs to maximize early exits across and greedily loading weights of most likely to be used layers, ~\ours{} reduces both token generation latencies and memory footprint. The memory savings lead to higher batch sizes and overall, \ours{} improves throughput by 45\%, unlike EE-LLMs that only improve it by 16\% relative to vanilla decoding.}
   \vspace{-0.20in}
   \label{fig:intro}
\end{figure}

\noindent \textbf{Limitations of Current EE-LLM Serving:} 
Existing EE-LLMs rely on a single model and tokens that do not meet the confidence threshold must traverse additional layers. This limits latency savings for tokens that cannot exit early.  
Moreover, EE-LLMs cannot increase batch sizes due to two reasons.
\textit{First}, they do not yield memory savings
because early exits taken depend on the request and are unknown apriori to serving it; and not all tokens exit early. Therefore, EE-LLMs load the weights of all layers on the GPUs and compute their Key-Value (KV) vectors to accommodate the worst-case exit. Figure~\ref{fig:intro} shows that the memory required to store the weights of two Llama models accounts for up to 68\% of available HBM even on state-of-the-art NVIDIA B100s and is identical for vanilla auto-regressive decoding and EE-LLMs.  Consequently, batch sizes cannot be increased because memory utilization remains unchanged. 


\textit{Second}, batched inference in EE-LLMs is non-trivial due to synchronization issues. Typically, in batching, a token is generated for every request, before proceeding to generate the next token for all requests in the batch. EE-LLMs must therefore, either wait for the token that takes the longest in each round or simply use a batch size of $1$. Given the synchronization overheads in the former, existing EE-LLMs use the latter by default~\cite{ee-llm, ee-tune}.

\begin{figure}[tp]
  \centering
  \includegraphics[width=\columnwidth]{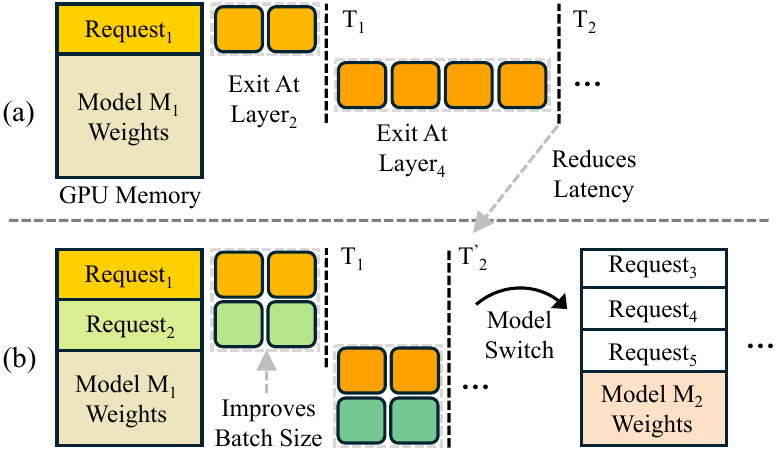}   
   \vspace{-0.25in}
    \caption{(a) Current EE-LLMs select a model (say $M_1$), load weights of all its layers, and use a batch size of $1$ to avoid synchronization across tokens. (b)~\ours{} uses multiple LLMs ($M_1$ and $M_2$ here) and only loads the weights of the layers most likely to be used based on real-time early exit profiles. \ours{} improves batch sizes by increasing available memory capacity and reducing synchronization overheads.~\ours{} also monitors performance in real-time and switches between LLMs or loads additional layers of the current model to prevent accuracy degradation. }
   \label{fig:intro_design} \vspace{-0.15in}
\end{figure}

\noindent \textbf{Our Proposal:} We propose {\em \ours}, a framework that improves both token generation latency and batch sizes to enable high-throughput in EE-LLMs. \ours{}, shown in Figure~\ref{fig:intro_design}, exploits two key insights. {\em First}, tokens that do not exit early on one LLM often exits early on another. For example, our studies show that the first six layers of the 24-layer OPT-1.3B model processes 74\% of tokens for a prompt mix of standard benchmarks, while the remaining 26\% require all layers. However, 57\% of these remaining tokens can be served by using only the first nine layers of the 32-layer OPT-6.7B model. 
Thus, \ours{} efficiently uses multiple LLMs to maximize the number of early exit tokens and lower the average token generation latencies drastically. For instance, by judiciously using both OPT-1.3B and 6.7B, \ours{} produces 92\% of the tokens using early exits, compared to only 74\% and 77\% respectively by using them standalone. Note that significantly fewer tokens (only 8\% in this case) now require additional layer traversal.

\textit{Second}, our studies show that even if the confidence threshold is not met at an early exit, the predicted token often remains \textit{unchanged} even after additional layer traversal. For example, our studies with CNN-Dailymail dataset~\cite{cnn-daily-mail} and OPT-6.7B model show that there is a 92.1\% chance that tokens prevented from exiting at Layer-9 due to a low confidence score of 0.2, ultimately becomes the final output token after traversing through all 32 layers. Thus, most tokens that do not meet the confidence threshold can still be greedily obtained from early exits; and weights corresponding to the later layers would then remain largely unused. \ours{} leverages this insight to only load the weights of the most likely to be used layers (such as up to Layers 6 and 9 for the OPT-1.3B and 6.7B models respectively in the above example). This yields memory savings (3.37GB and 17.25GB respectively in this case) which is repurposed to support larger batch sizes. Note that this approach eliminates synchronization overheads between tokens in a batch because it ensures that
each token only traverses a fixed number of model layers.

\noindent \textbf{Key Challenges of \ours{}:}
Employing multiple LLMs, dynamically switching between them, and maximizing early exits is non-trivial due to several reasons. \textit{First}, the early exits taken depend on the request and are unknown. \textit{Second}, 
frequent model switching incurs overheads and causes slowdown. \textit{Third}, aggressively allowing tokens to exit early when confidence is not met degrades accuracy. To address the first challenge, \ours{} employs real-time profiling to accurately identify the distribution of early exits for incoming requests. As successive requests often have overlapping contexts and exhibit locality~\cite{lin2024parrot}, this early exit profile remains relevant and accurate temporally. 

To minimize switching overheads and accuracy loss, \ours{} uses a two-step approach. \textit{First},
it tracks a window of consecutive tokens and detects potential accuracy degradation if a certain number of tokens fail to meet the confidence threshold. \textit{Second}, under such circumstances,  \ours evaluates two options-- either (1)~load all layers of the model currently in use or (2)~switch to an alternate model that can potentially complete the task more efficiently by using early exits. \ours{} evaluates the overheads of each option using the real-time telemetry data gathered during profiling and selects the option with minimal overheads.


Overall, this paper makes the following contributions:\vspace{-0.1in}
\begin{enumerate}[leftmargin=0cm,itemindent=.5cm,labelwidth=\itemindent,labelsep=0cm,align=left, itemsep=0 cm, listparindent=0cm]

\item We show that the throughput of EE-LLMs is limited because relying on a single LLM restricts computational savings for tokens that do not exit early, whereas the lack of memory savings limit us from scaling batch sizes.

\item We propose\textit{ \ours{}}, a framework  that uses multiple models and dynamically switches between them to collectively maximize the number of early-exit tokens. This reduces token generation latencies and increases throughput. 


\item~\ours exploits the observation that low-confidence tokens which could not exit early, often remain unchanged even after additional layer traversal and steers more tokens to exit early, increasing throughput even further.

\item \ours{} profiles the early exit distributions in real-time and greedily loads the weights of only the most likely to be used layers. The memory savings 
enable higher batch sizes.



\end{enumerate}

\vspace{-0.05in}
Our evaluations show that~\ours improves throughput and batch sizes by {$1.48\times$ and 15.14$\times$ respectively compared to current EE-LLMs, with negligible impact on accuracy.} 

\clearpage

\section{Background and Motivation}\label{sec:background}

\subsection{Early Exit Large Language Models}
\textit{Early-Exit LLMs} or \textit{EE-LLMs} are a class of language models that enable high throughput inference by allowing tokens to exit at specific intermediate layers during the forward pass if their probability meets a predefined confidence threshold. EE-LLMs reduce the average token generation latency by skipping computations in the later model layers for simple tokens and improve throughput without degrading accuracy. 
The computational and latency savings with EE-LLMs scale proportionally with the number of layers skipped which eventually translate into higher throughput.  Consequently, EE-LLMs have been widely adopted in various deep learning architectures in both industry and academia~\cite{elhoushi2024layerskip, wang2024early, xu2023lgvit}.

\subsection{Limitations of Existing EE-LLM Serving}

Existing EE-LLM frameworks yield limited throughput benefits due to two key limitations. \textit{First}, they yield \textit{limited latency savings} 
because they rely on a single model and tokens that fail to exit early on this model must traverse additional layers until the confidence threshold is met. Consequently, the average token generation latency and throughput are limited by the number of tokens that do not exit early.

\textit{Second}, EE-LLMs offer \textit{no memory savings} which limits batch sizes. As early exits taken are only known at runtime and cannot be predicted in advance, current EE-LLMs load weights of all the model layers on to the GPUs, even though the later layers remain unused when tokens exit early. Note that despite recent advances in quantization~\cite{zhao2024atom} and compression~\cite{zhu2024survey}, model weights dominate the GPU memory footprint. For example, the Llama3.1 405B model consumes about 52\% of the available HBM to store model weights even on a node with eight state-of-the-art NVIDIA B100 GPUs. Moreover, current EE-LLM frameworks also compute and cache the Key-Value (KV) vectors for all skipped layers to account for the worst-case exit depth. This is because any future token that does not exit early, must attend to all preceding tokens. Thus, overall, the memory footprint of EE-LLMs is identical to vanilla auto-regressive decoding, limiting batch sizes.


\subsection{Goal: Maximize Early Exit Tokens and Batch Size}

As token generation latency depends on the number of layers traversed, ideally, we want to maximize the number of tokens that exit early to maximize throughput. To improve throughput even further, we must scale to larger batch sizes by improving the memory-efficiency such that we reduce the footprint of unused memory and re-purpose the memory savings to support additional requests in parallel. This paper proposes \textit{\ours{}} that achieves these goals.

\section{Our Proposal: \ours{}}\label{sec:design}
In this paper, we propose {\em \ours{}}, a framework that improves the throughput of EE-LLMs by maximizing the number of early exit tokens, thereby reducing the average token generation latency,  and increasing batch sizes through efficient memory management. Next, we discuss the key insights of our design before describing the implementation.


\subsection{Key Insights}
{\em \ours} leverages two key insights that are described next. 

\begin{figure}[htp]
  \centering
  \includegraphics[width=1.0\columnwidth]{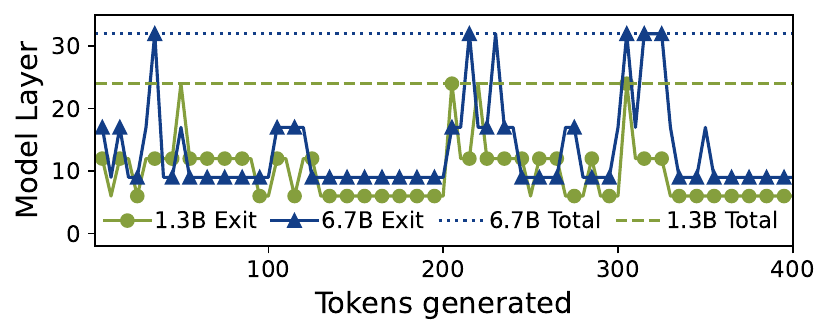}\vspace{-0.2in}
   \caption{Exit layers for serving a typical workload with a mixture of prompts using OPT 1.3B and 6.7B models.}
   \label{fig:exitselection}
\end{figure}

\textbf{\textit{Insight-1 $\rightarrow$ Early exits are often complimentary:}} Our experiments show that early exits taken depend on the EE-LLM and are often complimentary across models. Tokens that require additional layer traversal or all layers of a model (i.e. no early exits taken) can often be predicted accurately with another model using early exits. 
For example, Figure~\ref{fig:exitselection} shows that prompts needing more than twelve layers on the 24-layer OPT-1.3B model require fewer layers on OPT-6.7B for similar accuracy. We make similar observations for prompts needing more than nine layers on the 32-layer OPT-6.7B model. ~\ours{} leverages this characteristic to employ multiple models and dynamically switches between them such that both the models collectively maximize the number of early exit tokens, thereby reducing the average token generation latency and improving throughput.

\begin{figure*}[tp]
  \centering
  \includegraphics[width=\linewidth]{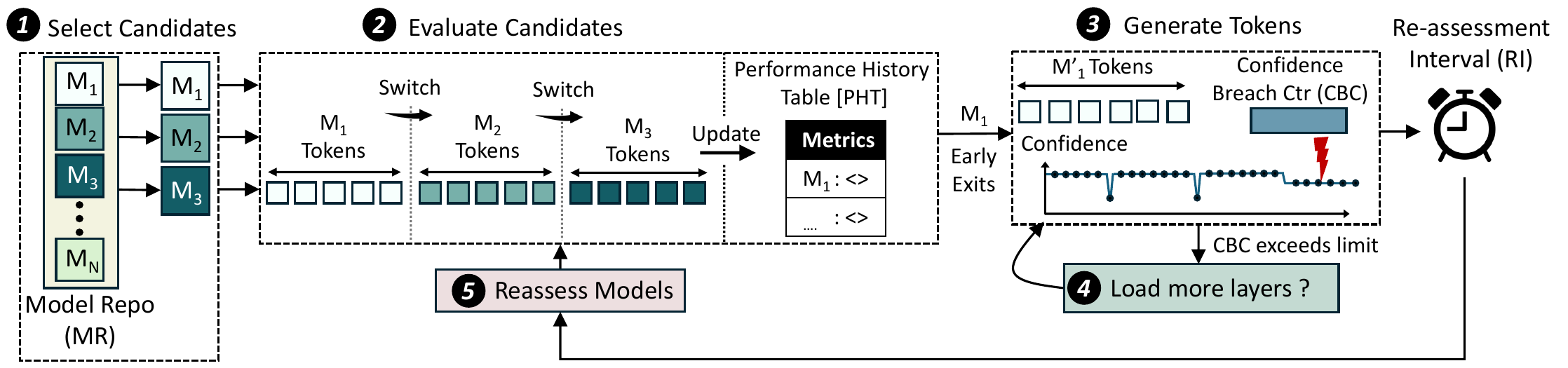}
   \caption{Design of \ours}  \vspace{-0.1in}
   \label{fig:designoverview}
\end{figure*}

\begin{figure}[htp]
   \centering
   \includegraphics[width=0.8\columnwidth]{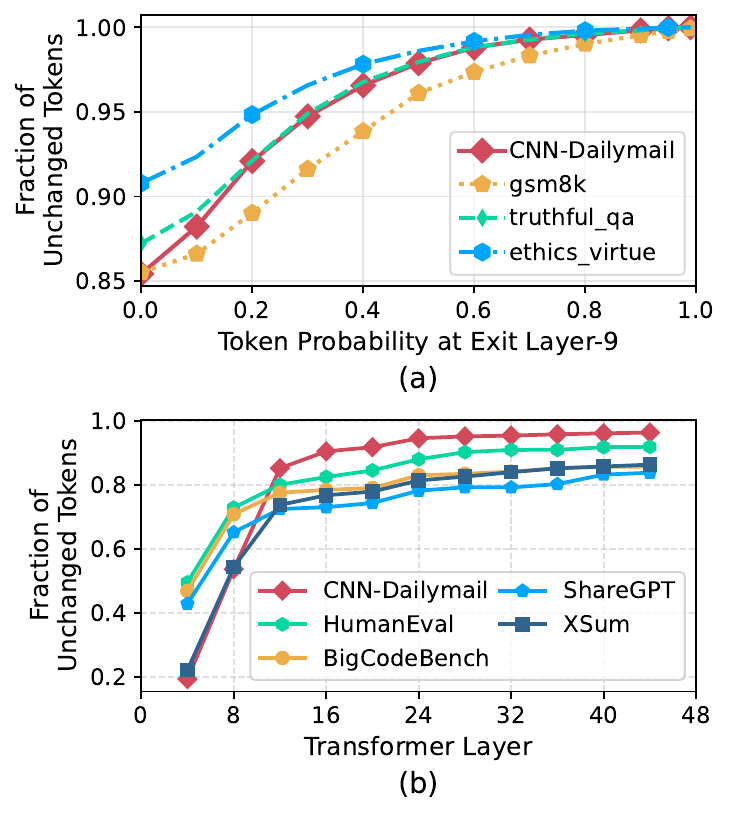}
   \vspace{-0.25in}
   \caption{ (a) Fraction of unchanged tokens for four datasets on OPT-6.7B model from the 1st exit layer (9) to the final layer (32). We observe that probability of the predicted token staying unchanged is always greater than 85\%. (b) Fraction of tokens generated at an exit-layer that remain unchanged even after traversing the full-model for the Codellama-34B model across six datasets. }
   \vspace{-0.2in}
   \label{fig:llama_exit_dist}
\end{figure}

\textbf{\textit{Insight-2 $\rightarrow$ Not meeting confidence is okay at times:}} 
Our studies reveal that even when the confidence threshold is not met at an early-exit, the predicted token frequently remains \textit{unchanged}, even after traversing additional model layers. Figure~\ref{fig:llama_exit_dist}(a) shows the fraction of tokens that remain unchanged even after traversing additional model layers of the 32-layer OPT 6.7B model as a function of their probability at the first early exit (Layer-9). 
For instance, if the confidence threshold is 1, no token would exit early. However, we observe that even if we consider the threshold to be as low as 0, the token predicted at Layer-9 is identical to the token predicted at the final exit Layer-32 for 85\% of the cases; and traversing through subsequent layers only improves the confidence. A prior work~\cite{bert-patience} also makes a similar observation. 

Furthermore, early-exits consistently produce accurate output tokens even across different model sizes and families. For example, Figure~\ref{fig:llama_exit_dist}(b) shows that even for Codellama-34B, 90\% of the tokens produced by Layer-16 (one-third the model depth) remain unchanged for the CNN-Dailymail~\cite{cnn-daily-mail} dataset (more data provided in Appendix~\ref{app:early_good}).  
\ours{} leverages this insight to greedily load only the most likely to be used layers of the selected model and frequently allows tokens to exit even if the confidence threshold is not met, yielding memory savings that enable larger batch sizes. Note that the impact of this on accuracy is negligible because (1)~\ours{} already drastically reduces the percentage of tokens requiring additional layer traversal by employing multiple models with complimentary early-exit characteristics and (2) not meeting the confidence threshold does not mean the predicted token is incorrect. \ours{} also introduces additional steps in the design to minimize accuracy loss, as will be discussed next. 

\subsection{Design Overview}
Figure~\ref{fig:designoverview} shows an overview of \ours. \bcircled{1} \ours selects a set of candidate models and \bcircled{2} evaluates their performance in real-time. \bcircled{3} The chosen model is then only loaded up to a limited number of layers based on the early exit history from the evaluation step and is used to generate tokens further. \bcircled{4} If the number of layers loaded for the current model are insufficient, the system requests for additional layers. \ours compares the overheads of (1) loading more layers for the current model versus (2) switching to another model from the candidate pool, and decides on one of them depending on their overheads. \bcircled{5} \ours also periodically reassesses the performance of the selected model and switches to another candidate model if needed, to adapt to the changing characteristics of the request stream.

\subsection{Design Implementation}
Next, we discuss the implementation of \ours. 

\subsubsection{Step-1: Selection of Candidate Models}

Typically, service providers maintain a \textit{Model Repository (MR)} containing key performance metrics, such as throughput, accuracy, across different standard benchmarks. \ours uses this telemetry data to select the \textit{TopK} candidate models based on user-specified SLOs and available hardware.
We illustrate this in Figure~\ref{fig:designoverview} using models $M_1$, $M_2$, and $M_3$ chosen as candidates. By default, \ours selects up to three models to minimize the time spent on the evaluation step, which is described next, and quickly converge on the suitable candidate for the current request stream.

\subsubsection{Step-2: Evaluation of Candidate Models}
Next, \ours evaluates the performance of the selected candidate models in real-time to obtain even more accurate telemetry data and gather the early-exit distribution profiles because they are not present in model repositories by default. This profile remains effective because successive queries often share \textit{overlapping contexts} and exhibit locality. 
By default, \ours evaluates each candidate model for five requests. Figure~\ref{fig:designoverview} illustrates this, where models $M_1$, $M_2$, and $M_3$ generates output tokens for five requests. This does not impact performance (such as time taken to generate first token or TTFT) because the output tokens generated are not discarded given that the preceding candidate selection process is highly selective and ensures that only competent models are shortlisted. Note that candidate model evaluation may also be done in parallel, but we avoid it in our default design due to hardware limitations (not enough GPUs).



\noindent \textbf{Methodology:} We obtain throughput and early-exits distribution using profiling tools. Assessing accuracy is non-trivial because incoming requests lack ground truth for comparison. So, we use \textit{perplexity} because it is a reference-free, efficient, and suitable metric for real-time inference (details in Appendix~\ref{app:perplexity}). Table~\ref{tab:evalimpact} shows the perplexity of two models before and after evaluation and highlights the efficacy of our approach. If we were to select a model based on the repository data, we would select the OPT-6.7B model. However, we select the OPT-1.3B model because post-evaluation data suggests it to be more effective for the current prompts. 





\begin{table}[htb]
    \centering
    \vspace{-0.1in}
    \begin{center}
    \caption{Perplexity comparison between pre-determined value from MR in selection phase and post evaluation phase.} \vspace{0.05in}
    \label{tab:evalimpact}
    \setlength{\tabcolsep}{5pt}
    \renewcommand{\arraystretch}{1.2}
    \begin{tabular}{|c|c|c|}
        \hline
        Model & Pre-Evaluation & Post-Evaluation \\
        \hline
        \hline
        OPT-1.3B & 1.91 & 1.47 \greentick \\
        \hline
        OPT-6.7B & 1.68 \greentick & 1.49 \\
        \hline
    \end{tabular}
    \vspace{-0.15in}
    \end{center}
\end{table}

\noindent \textbf{Performance History Table:} The request-specific performance data (throughput, accuracy) and the early exit profiles of each model are saved in a table, called the \textit{Performance History Table (PHT)}. The PHT is used in the subsequent stages of \ours, as will be discussed next.


\subsubsection{Step-3: Token Generation Using Best Model} 
\noindent The best candidate model identified in the evaluation stage is used to generate tokens for the incoming requests. 

\begin{figure}[htp]
\centering
\includegraphics[width=1.0\columnwidth]{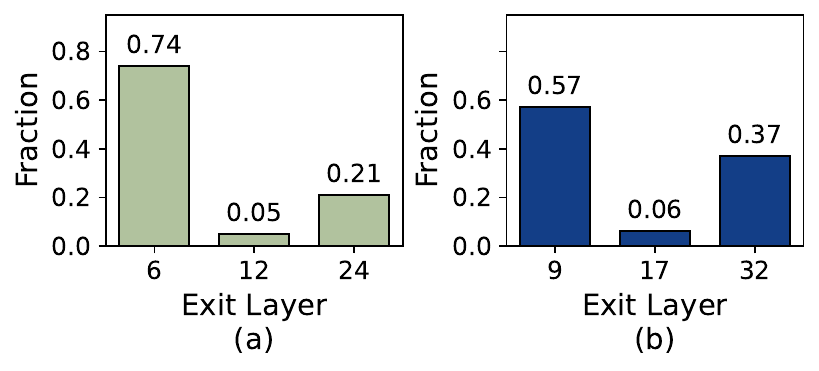}%
\vspace{-0.10in}
\caption{Distribution of exit layers for tokens of a prompt mix using the OPT-1.3B Model. 74\% of the requests only require up to 6 layers. (b) Distribution of exit layers of the remaining 26\% when they are serviced by OPT-6.7B model.}%
\label{fig:histogram}%
\vspace{-0.2in}
\end{figure}

\vspace{0.05in}
\noindent \textbf{Greedy Loading Up to Selected Exit Layers:} 
\ours greedily loads the weights of the most likely to be used layers based on the early exit profile of the model. This yields memory savings which is used to increase batch sizes \textbf{\textit{(Insight-\#2)}}. 
For example, Figure~\ref{fig:histogram}(a) shows the exits taken for a prompt mix with the OPT-1.3B model, where 74\% of the requests only require six layers of the model. We refer to these as \textit{Low-Exit Tokens (LTs)}. \ours greedily loads only up to six layers in this scenario (denoted by $M_1'$ in Figure~\ref{fig:designoverview}). If most pending requests are LTs that do not require additional layers, this greedy approach yields significant memory savings, without compromising accuracy.


\vspace{0.05in}
\noindent \textbf{Load More Layers Or Another Candidate Model?} 
\noindent Although partially loaded models offer significant memory savings and improve batch sizes, we encounter tokens that do not meet the confidence threshold. For example, in Figure~\ref{fig:histogram}(a), 26\% of the requests use more than six layers. We refer to these as \textit{High-Exit Tokens (HTs)}. 
Under these circumstances, \ours has two options-- (1)~either load the remaining layers of the current LLM (OPT-1.3B in this case) or (2)~switch to another model which can service the request stream with fewer layers. The first option ensures the current model is present in its entirety and guarantees that the confidence threshold will be met. In contrast, the second option aims to identify a more efficient alternative and maximize the total number of early-exits (\textit{\textbf{Insight-\#1}}). Figure~\ref{fig:histogram}(b) shows the distribution of the exits taken by the HTs when another candidate model, OPT-6.7B, is used. We observe that 57\% of the HTs can be serviced by using only nine layers of the OPT-6.7B model. Note that the exit history information is already available in the PHT from Step-2. Loading more layers of the current model is beneficial only when the overall resource usage remains lower than the second option where another candidate model is loaded up to a limited number of layers (OPT-6.7B with nine layers in the example). Once the overheads of both options are evaluated, the option with minimal overheads is selected and appropriate action is taken.

\vspace{0.05in}
\noindent \textbf{Amortizing Loading Overheads with CBC:} 
Irrespective of the option selected, both approaches-- loading additional layers and model switching, incur overheads. 
To minimize these overheads, \ours 
considers both options only after a certain number of tokens within a window fail to meet the confidence threshold. This leverages our observation that not every token that fails to meet the confidence threshold at an early exit, actually changes after future layers (\textit{\textbf{Insight-\#2}}). ~\ours{} uses a \textit{Confidence Breach Counter (CBC)} that increments whenever an output token does not meet the confidence threshold. If the CBC exceeds a pre-determined maximum allowable limit (CBC$_{max}$),~\ours{} re-assesses whether it should load more layers or switch to another model to serve the incoming request stream. By default, \ours only tolerates up to 50 confidence threshold breaches (CBC$_{max}=50$) in a window of 100 consecutive tokens.



\subsubsection{Step-4: Periodic Profiling to Maximize Early-Exits} 

\begin{figure}[h]
  \centering
  \includegraphics[width=\columnwidth]{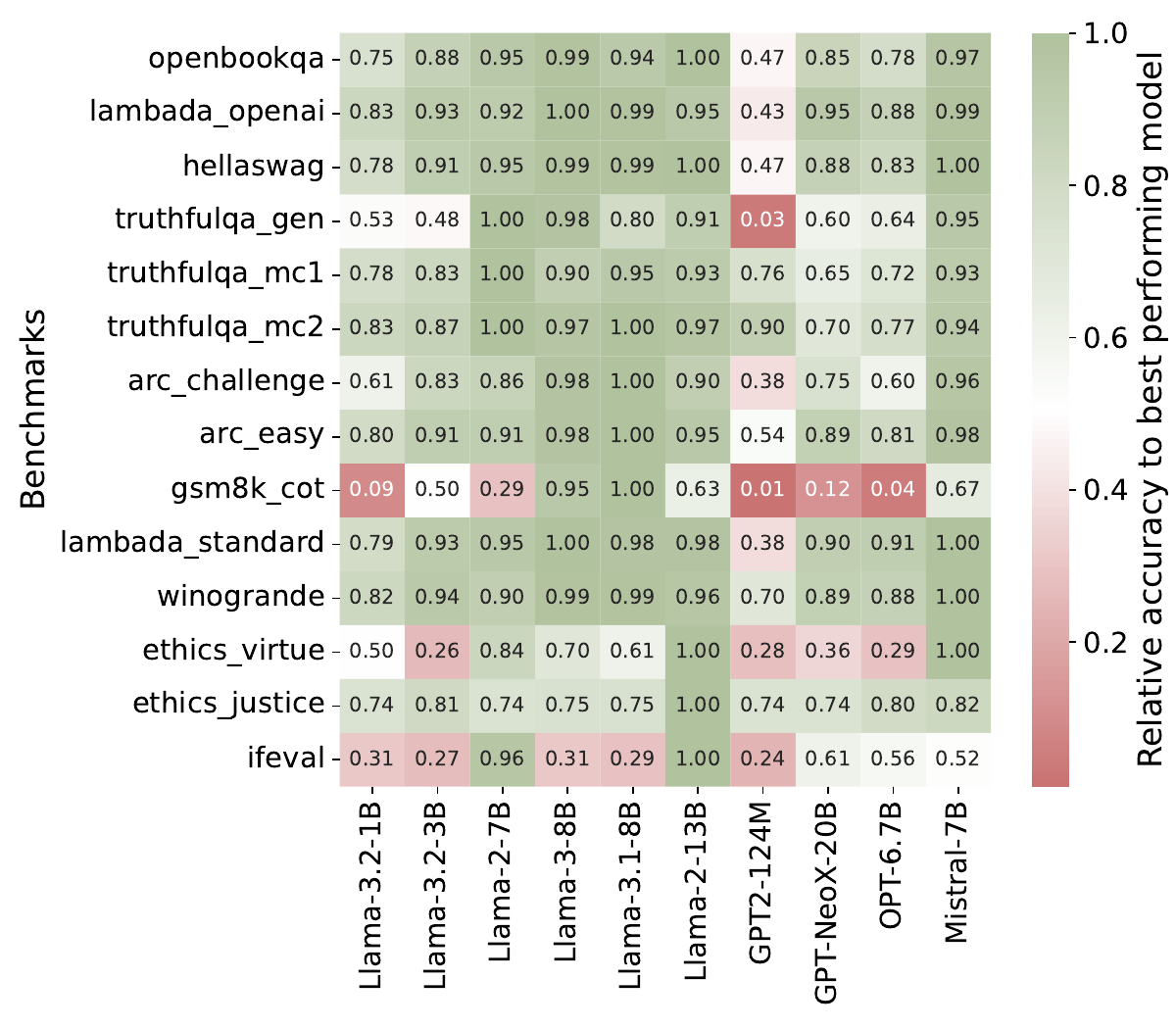}
   \caption{Accuracy of datasets across multiple models relative to the best  model. We observe that some models consistently perform well (such as Llama3-8B, Llama2-13B) while some others (such as GPT2-124M) remain consistently poor. \ours{} exploits this observation to eliminate frequent candidate model selection and only performs this step when SLOs or hardware constraints change.}  
   \label{fig:heatmap}
\end{figure}

Ideally, we should re-evaluate early exit profiles for each request across all candidate models to maximize early-exits for the current request stream. However, frequent profiling introduces substantial overheads. In contrast, capturing the early-exit profile infrequently is not desirable because although input requests exhibit temporal locality, their characteristics often evolve over longer time periods. Consequently, the early exit profile captured in the PHT that is currently being used by~\ours{} may become obsolete and sub-optimal. \ours{} attains a sweet spot in this trade-off space by using the \textit{Re-assessment Interval ($RI$)} hyperparameter and invokes the profiling phase (Step-2) periodically after $RI$ requests have been served. By default, \ours{} collects early-exit profiles after every 150 requests ($RI$=150) but ideally, this hyper-parameter must be fine-tuned by the server provider to match the variations in their request streams. Our default implementation does not reselect candidate models because our evaluations show that some models generally perform better across a wide variety of tasks. For example, Figure~\ref{fig:heatmap} shows that Llama3-8B and Llama2-13B consistently perform well, whereas GPT2-124M is consistently poor. Nonetheless,~\ours can also look up the model repository to select new candidate models, if the user-specified SLOs or hardware constraints change.

\subsection{~\ours{} Algorithm}

Algorithm~\ref{alg:helios} describes the algorithm of \ours.

\begin{algorithm}
\caption{Adaptive EE-LLM Serving with \ours}
\label{alg:helios}
\begin{algorithmic}
   \STATE {\bfseries Input:} Model Repository(MR), SLO
   \STATE {\bfseries Output:} Dynamic Model Selection
   \STATE {\bfseries Parameters:} $\mathbf{M}$: Full Model; $\mathbf{M'}$: Low Exit Model; $\mathbf{CBC}$: Confidence Breach Counter; $\mathbf{CBC}_{\max}$: Threshold; $\mathbf{RI}$: Reassessment Interval; 
   \STATE \textcolor{blue}{Step-1:} Candidates$\gets$$Top_k$(MR(SLO, HW constraints))
   \WHILE{prompts in requests}
      \STATE $\mathbf{CBC},$ ServicedPrompts $\gets 0$
      \STATE \textcolor{blue}{Step-2:} PHT$[M,M'] \gets$ Evaluate(Candidates)
      \STATE Chosen $\gets$ BestModel(PHT[$M'$])
      \REPEAT 
         \STATE \textcolor{blue}{Step-3:} Serve(prompt, Chosen)
         \IF{confidence not met}
            \STATE CBC $\gets$ CBC + 1
            \IF{CBC $>$ CBC$_{\max}$}
               \IF{PHT[$M$(Chosen)] $<$ PHT[$M'$(Others)]}
                  \STATE Chosen $\gets M[$Chosen$]$
               \ELSE
                  \STATE Chosen $\gets M'[$NextBestModel(PHT)$]$
               \ENDIF
               \STATE CBC $\gets 0$
            \ENDIF
         \ENDIF
      \UNTIL{ServicedPrompts $<$ RI (\textcolor{blue}{Step-4})}
   \ENDWHILE
\end{algorithmic}
\end{algorithm}

\newpage
\section{Evaluation Methodology}
We discuss the methodology used to evaluate \ours.

\subsection{Models}
We use publicly available early-exit variants of the Llama models hosted on HuggingFace~\cite{elhoushi2024layerskip}. Additionally, we augment and fine-tune two off-the-shelf models from Facebook's OPT family~\cite{zhang2022opt}, namely OPT-1.3B and OPT-6.7B, by incorporating early-exits at one-fourth the model depth. This strategy is consistent with prior works~\cite{ee-llm}. We provide more details on models and fine-tuning in Appendix~\ref{app:models}.
Table~\ref{tab:model_configs} summarizes the candidate model configurations used for our evaluations. By choosing diverse model families and sizes, our evaluations ensure that we capture diverse scenarios and shows the generalizabilty of our approach.

\begin{table}[!htb]
    \centering
    \begin{center}
    \caption{Overview of Candidate Model Configurations}
    \label{tab:model_configs}
    \renewcommand{\arraystretch}{1.2}
    \begin{tabular}{|c|l|}
        \hline
        Configuration & Candidate Models\\
        \hline
        \hline
         1 & OPT-1.3B \& OPT-6.7B\\
        \hline
         2 & Llama2-7B, 13B \& Llama3-8B\\
        \hline
         3 & CodeLlama-34B \& Llama2-70B\\
        \hline
    \end{tabular}
    \end{center}
\end{table}

\ignore{
We consider two models of varying sizes from Facebook's OPT family~\cite{zhang2022opt} with 1.3 and 6.7 billion parameters. We add early exits to these models at 1/4th the depth~\cite{ee-tune, ee-llm}. Prior work EE-Tuning~\cite{ee-tune} notes that exits at shallower depths, such as 1/4th the model depth, enable faster inference while maintaining performance. Table~\ref{tab:models} summarizes the models. The open-source implementations of the models in Table~\ref{tab:models} does not include early exits. So, we fine-tune them to calibrate the early exits. To ensure generalization across tasks, we use the red-pajama~\cite{redpajama} and pile~\cite{the-pile} datasets. For fine-tuning, we use a weight of 1.0 to each early-exit loss in the total loss calculation and use 50,000 iterations.

\begin{table}[!htb]
    \centering
    \begin{center}
    \caption{Summary of Models Used} 
    \label{tab:models}
    \renewcommand{\arraystretch}{1.2}
    \begin{tabular}{|c|c|c|c|}
        \hline
        Model & Parameters & Layers & Early Exits \\
        \hline
        \hline
        OPT-1.3B & 1.3 billion & 24 & 6, 12, 24\\
        \hline
        OPT-6.7B & 6.7 billion & 32 & 9, 17, 32\\
        \hline
    \end{tabular}
    \vspace{-0.15in}
    \end{center}
\end{table}
}

\subsection{Setup}
We use the EE-LLM framework~\cite{ee-llm} for both fine-tuning and inference serving, which is consistent with prior works~\cite{ee-llm,ee-tune}. We limit our evaluations to three models due to limited access to GPUs. We use a node equipped with four NVIDIA A100 (40 GB) GPUs and 64-core AMD EPYC CPU. The GPUs are inter-connected via an all-to-all NVLink, providing upto 400 GB/s bidirectional bandwidth. Due to the large model weights of Llama2-70B and CodeLlama-34B, tensor parallelism of 4 and 2 is applied for these models, respectively, whereas all other models use a tensor parallelism of unity.

\begin{figure*}[tp]
  \centering
  \includegraphics[width=\linewidth]{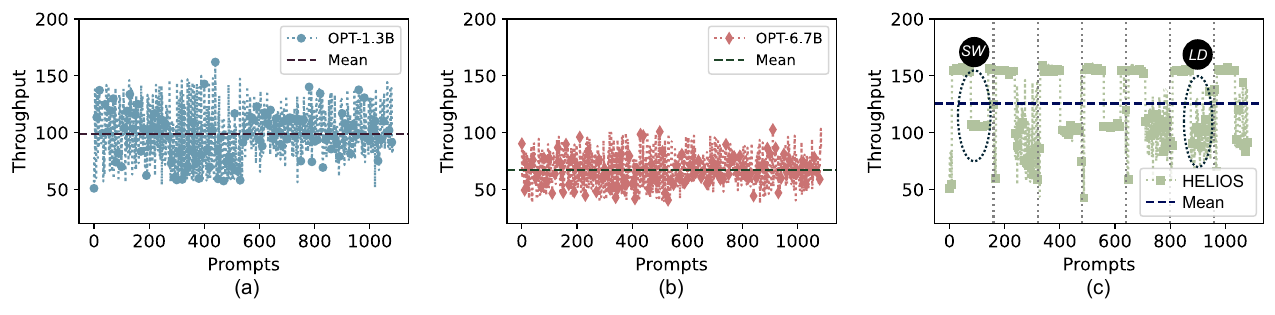}
   \caption{Comparison of \textit{Throughput} when using (a) only OPT-1.3B (EE-LLM), (b) only OPT-6.7B (EE-LLM), and (c) \ours (\textit{higher is bette}r). Candidate re-assessment steps are denoted using vertical dotted lines in (c).}
   \vspace{-0.1in}
   \label{fig:throughput}
\end{figure*}

\subsection{Datasets}
To evaluate the efficacy of~\ours{} at a server scale, we use the ShareGPT~\cite{sharegpt} dataset which comprises multi-turn user-LLM interactions. ShareGPT is a representative workload for analyzing system performance at server-scale as it is directly sourced from a live inference server. We further evaluate~\ours{} on a request stream composed of a wide range of tasks. This mixture draws requests from standard benchmarks including CNN Daily Mail~\cite{cnn-daily-mail}, gsm8k\cite{gsm8k}, CodeXGLUE~\cite{codexglue} and HellaSwag~\cite{hellaswag}. We provide additional details about these datasets in Appendix~\ref{app:benchmarks}.


\subsection{Figure-of-Merit}
We use perplexity to evaluate the accuracy of the generated tokens. This is consistent with prior works~\cite{federici2024efficient,frantar2023sparsegpt,xu2024besa,ma2023llm} (additional details in Appendix~\ref{app:perplexity}). To assess server performance, we use other widely used metrics such as Time Taken to First Token (TTFT), Time Per Output Token (TPOT), latency, and throughput, as summarized in Table~\ref{tab:metrics}. 
Our primary evaluation of~\ours assumes \textit{improving throughput} as the user-specified SLO. But we also consider other SLOs and provide experimental results in Appendix~\ref{sec:otherslos} to show that \ours{} is generalizable.

\begin{table}[!htb]
    \centering
    \begin{center}
    \caption{Summary of Metrics Used} 
    \label{tab:metrics}
    \renewcommand{\arraystretch}{1.1}
    \begin{tabular}{|c|c|}
        \hline
        Metric &  Specification\\
        \hline
        \hline
        Perplexity & Captures coherence in output tokens \\
        \hline
        TTFT & Time Taken to First Token\\
        \hline
        TPOT & Time Taken Per Output Token\\
        \hline
        Latency & TTFT + TPOT $\times$ Number of Tokens\\
        \hline
        Throughput & Tokens generated per second ($\frac{1}{TPOT}$) \\
        \hline
        Batch Size & Number of requests served in parallel \\
        \hline
    \end{tabular}
    \vspace{-0.1in}
    \end{center}
\end{table}
\section{Results}
\label{sec:results}
By default, we consider the user's SLO is to maximize the inference throughput.

\subsection{Throughput}
For this particular evaluation, we restrict \ours to a batch size of 1 because it allows us to thoroughly evaluate the throughput benefits that solely stem from maximizing early exits. We also consider a prompt mix of standard benchmarks to evaluate the generalizability. The throughput is inversely proportional to the number of layers traversed and time spent per layer. Thus, it increases if more tokens (1)~take early exits and (2)~we use smaller models for early exits because traversing a layer takes longer on larger models. Figure~\ref{fig:throughput} shows the \textit{throughput (higher is better)}. \ours improves the throughput by $1.48\times$ and $2.13\times$ on average compared to using the OPT-1.3B and OPT-6.7B models standalone EE-LLMs respectively. In Figure~\ref{fig:throughput}(c), {\em LD} illustrates scenarios where more layers of the current model are loaded, whereas {\em SW} denotes cases where \ours switches to another model, highlighting the efficacy of the adaptive nature of \ours. 

\begin{figure*}[tp]
  \centering
  \includegraphics[width=\linewidth]{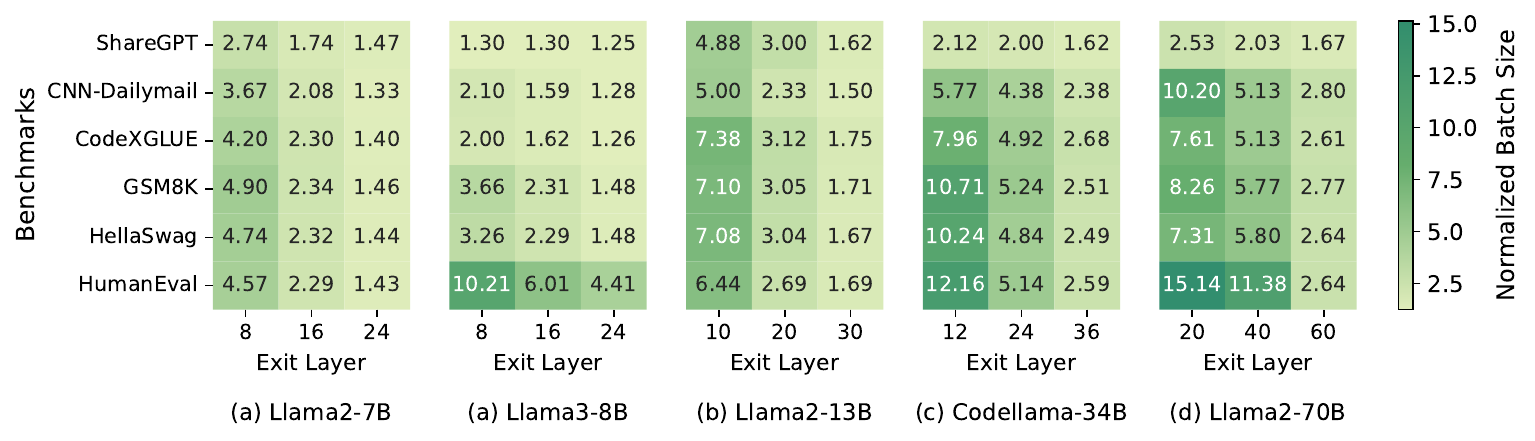}
  \vspace{-0.3in}
  \caption{Normalized batch size with~\ours{} compared to current EE-LLM framework. Greedily loading only most-likely to be used layers reduces the footprint of both the weights and Key-Value (KV) caches, yielding significant memory savings which are repurposed to support other requests in parallel and increasing the batch sizes.}  
  \label{fig:llama_batch}
\end{figure*}

Table~\ref{tab:exitdistribution} compares the exit distribution of tokens when the OPT EE-LLMs are used standalone against \ours. In \ours, about 91\% of the tokens are processed using the earliest exits of both models combined (Layer-6 of the OPT-1.3B model and Layer-9 of the OPT-6.7B model), compared to only about 73\% while using the models standalone. Also, a significant portion (about 70\%) of these tokens are processed using the earliest exit of the smaller OPT-1.3B model. The percentage of requests that use all layers of both models combined is only 7.39\%, $3\times$ lower than using either EE-LLM standalone. This highlights the efficacy of \ours in maximizing the number of early exit tokens and ensuring that the accuracy impact is minimal because tokens that require additional layers are not compromised.

\begin{table}[!htb]
    \centering
    \begin{center}
    \caption{Comparison of the percentage of tokens processed by different exit layers for different model selection methods.} \vspace{0.1in}
    \label{tab:exitdistribution}
    \setlength{\tabcolsep}{2pt}
    \renewcommand{\arraystretch}{1.1}
    \begin{tabular}{|c|c|c|c|c|c|c|c|}
        \hline
        \multirow{2}{*}{Model Selection}& \multicolumn{3}{c|}{OPT-1.3B} & \multicolumn{3}{c|}{OPT-6.7B}\\
        \cline{2-7}
         & 6 & 12 & 24 & 9 & 17 & 32 \\
        \hline
        \hline
        OPT-1.3B Only & 73.0 & 4.70 & 22.3 & - & - & -\\
        \hline
        OPT-6.7B Only & - & - & - &  73.6 & 4.80 & 21.6 \\
        \hline
        \ours & 70.19 & 1.38 & 6.78 &  20.90 & 0.14 & 0.61 \\
        
        \ignore{
        \hline
        \hline
        \multicolumn{7}{|c|}{\textit{SLO: Response Time Optimization}}\\
        \hline
        \ours & 49.3 & 0.20 & 0.90 & 48.9 & 0.10 & 0.6\\
        \hline
        \hline
        \multicolumn{7}{|c|}{\textit{SLO: Accuracy Optimization}}\\
        \hline
        \ours & 52.3 & 1.50 & 7.30 & 34.6 & 0.70 & 3.60\\
        \hline
        \hline
        \multicolumn{7}{|c|}{\textit{SLO: Energy Optimization}}\\
        \hline
        \ours & 74.3 & 1.80 & 7.90 & 15.3 & 0.10 & 0.6\\}
        \hline\end{tabular}
    \end{center}
\end{table}

\subsection{Accuracy}

\ours{} has negligible impact on accuracy compared to existing EE-LLMs. Our experiments show that the achieved perplexity in \ours for the prompt mix is only $0.01$ higher than the OPT-1.3B model. Note that for this particular scenario the OPT-1.3B model is more accurate than the OPT-6.7B model and is hence used for comparison.

\subsection{Response Time and Latency}

Table~\ref{tab:responsetime} compares the Time To First Token (TTFT) and Time Per Ouput Token (TPOT) for \ours against each EE-LLM standalone. We observe that the TTFT of \ours is significantly lower because by default it serves most requests using a smaller model with early exits.  Similarly, the token generation latency or TPOT is up to 46.6\% lower because \ours maximizes the total number of early-exit tokens.

\begin{table}[!htb]
    \centering
    \begin{center}

    \caption{Comparison of response time and latency for the standalone EE-LLM models and \ours.} \vspace{0.1in}
    \label{tab:responsetime}
    \renewcommand{\arraystretch}{1.05}
    \begin{tabular}{|l|c|c|}
        \hline
        Model & TTFT (s) & TPOT (s)  \\
        \hline
        \hline
        OPT-1.3B Only & 0.042 &  0.010 \\
        \hline
        OPT-6.7B Only & 0.069 & 0.015 \\
        \hline
        HELIOS & 0.033 & 0.008 \\
        \hline
    \end{tabular}
    \end{center}
\end{table}

\subsection{Batch Sizes}
\ours supports larger batch sizes by (1)~eliminating synchronization overheads and (2)~re-purposing the GPU memory saved via greedy loading of weights of only the most likely to be used layers. In~\ours, all tokens at any given timestep must exit the same early exit layer (up to which the selected model is loaded), thus completely eliminating the need for synchronization. Processing a request requires memory to store (1)~model weights, and (2)~key-value (KV) caches for each layer of the model to generate tokens. While model weights can be shared, each request must maintain its own KV-caches. Loading only a subset of layers yields substantial memory savings, as the memory footprint of both the model weights and KV cache decreases in proportion to the number of layers skipped from loading. This yields considerable memory savings, which is repurposed by~\ours{} to allocate KV cache space for additional requests. This enables \ours to support larger batch sizes. 

Figure~\ref{fig:llama_batch} compares the batch sizes with~\ours{} against current EE-LLM frameworks. Here, we consider additional benchmarks beyond the prompt mix considered earlier. \ours improves batch sizes by up to 15.14$\times$, highlighting its efficacy. Table~\ref{tab:memorysavings} compares the memory footprint of \ours against each EE-LLM standalone. \ours yields significant memory savings (up to 67.4\%). Note that for the scenario where CodeLlama-34B and Llama2-70B are used, \ours is not required to load the entire model at all.

Typically, \ours yields greater memory savings for larger models comprising more early exits. This is because weights occupy a substantial amount of GPU memory for large models. For example, the Llama2-70B model occupies 81.6\% of the available memory in our setup with 160 GB. Moreover, larger models offer greater flexibility for early exits. For example, fine-tuning early-exits uniformly gives nine early exit paths in the Llama2-70B model, compared to only three in the Llama2-7B model. 


\begin{table}[!htb]
    \centering
    \begin{center}

    \caption{Memory footprint on Nvidia A100 GPUs for individual models standalone and \ours for the ShareGPT dataset.} \vspace{0.1in}
    \label{tab:memorysavings}
    \renewcommand{\arraystretch}{1.05}
    \begin{tabular}{|l|c|}
        \hline
        Model & Weights Memory Size (GB) \\
        \hline
        \hline

        \multicolumn{2}{|c|}{\textit{Using only two candidate models}} \\
        \hline
        \hline
        CodeLlama-34B Only & 63  \\
        \hline
        Llama2-70B Only & 129  \\
        \hline
        HELIOS & 42 \\
        \hline
        \hline

        \multicolumn{2}{|c|}{\textit{Using three candidate models}} \\
        \hline
        Llama2-7B Only & 12.9 \\
        \hline
        Llama3-8B Only & 15.5  \\
        \hline
        Llama2-13B Only & 24.8  \\
        \hline 
        HELIOS & 11.5 \\
        \hline
    \end{tabular}
    \end{center}
\end{table}

Note that the memory savings in \ours are orthogonal to other memory optimization methods that reduce the size of the weights memory through quantization~\cite{zhao2024atom, liu2024spinquant} and KV caches~\cite{li2024snapkv, ghadia2025dialogue, zhang2023h2o, xiao2023efficient} independently. \ours can be combined with these approaches for even greater memory savings and higher throughput benefits.

\subsection{Scalability Across Confidence Thresholds}

\begin{figure}[htp]
  \centering
  \includegraphics[width=0.7\columnwidth]{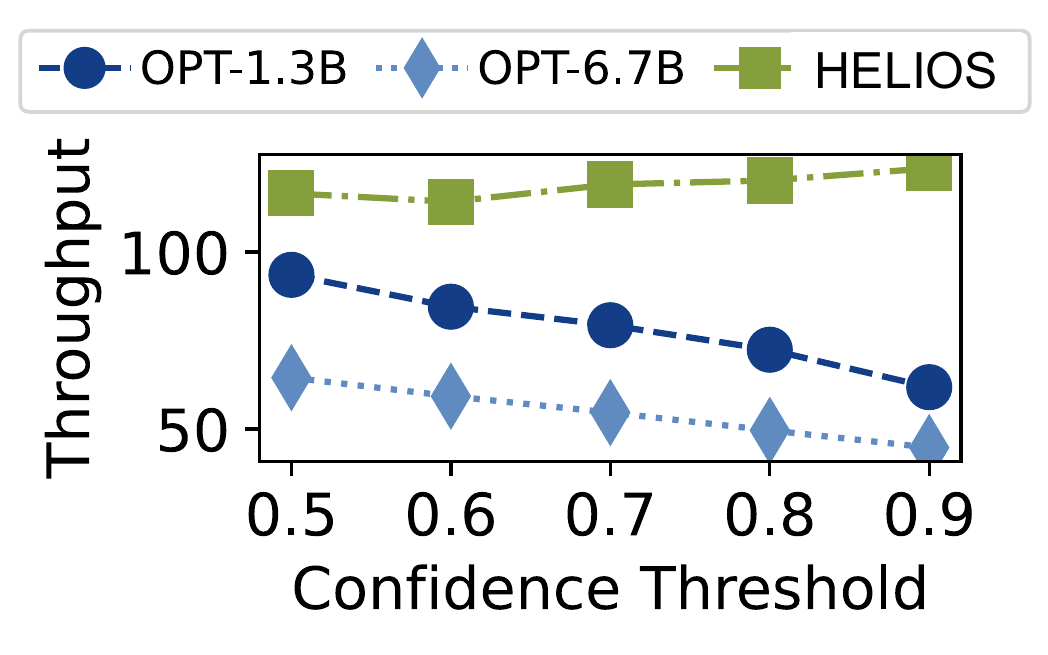}
   \vspace{-0.2in}
   \caption{Impact of increasing $TH$ on throughput.}\vspace{-0.1in}
   \label{fig:thresh_sens}
\end{figure}

The predefined confidence threshold ($TH$) for early exiting is a critical parameter that dictates the performance of EE-LLMs and \ours. Increasing the confidence threshold reduces the number of tokens taking early exits in existing EE-LLM serving frameworks because it becomes much harder to meet the exit criterion. Consequently, more tokens traverse additional layers, eventually reducing the effective throughput. This is consistent with our observation in  Figure~\ref{fig:thresh_sens} which shows the impact of increasing confidence thresholds on the throughput for two different EE-LLMs (OPT-1.3B and OPT-6.7B).

In contrast, the throughput benefits of \ours remains consistent. This is due to two reasons. \textit{First}, by employing multiple models, \ours encounters more tokens that naturally meet the confidence threshold and exits. \textit{Second}, by 
greedily loading weights of only those layers that are most likely to be used and enforcing tokens to breach the confidence thresholds, the impact of increasing confidence thresholds on throughput is minimized in \ours.


\subsection{Impact of Re-assessment}

\ignore{
    \begin{table}[!htb]
        \centering
        \begin{center}
        \caption{Specification of Design Hyper-parameters} 
        \label{tab:hyperparams}
        \renewcommand{\arraystretch}{1.1}
        \begin{tabular}{|c|c|}
            \hline
            Parameter &  Description\\
            \hline
                        \hline
            Confidence &  Minimum Probability with which an output  \\ 
            Threshold & token must be predicted for it to exit early\\
            \hline
            Re-assessment & Time taken between two consecutive \\
            Interval & candidate evaluation phases \\
            \hline

        \end{tabular}
        \end{center}
    \end{table}

}

\begin{figure}[htp]
  \centering
  \includegraphics[width=0.7\columnwidth]{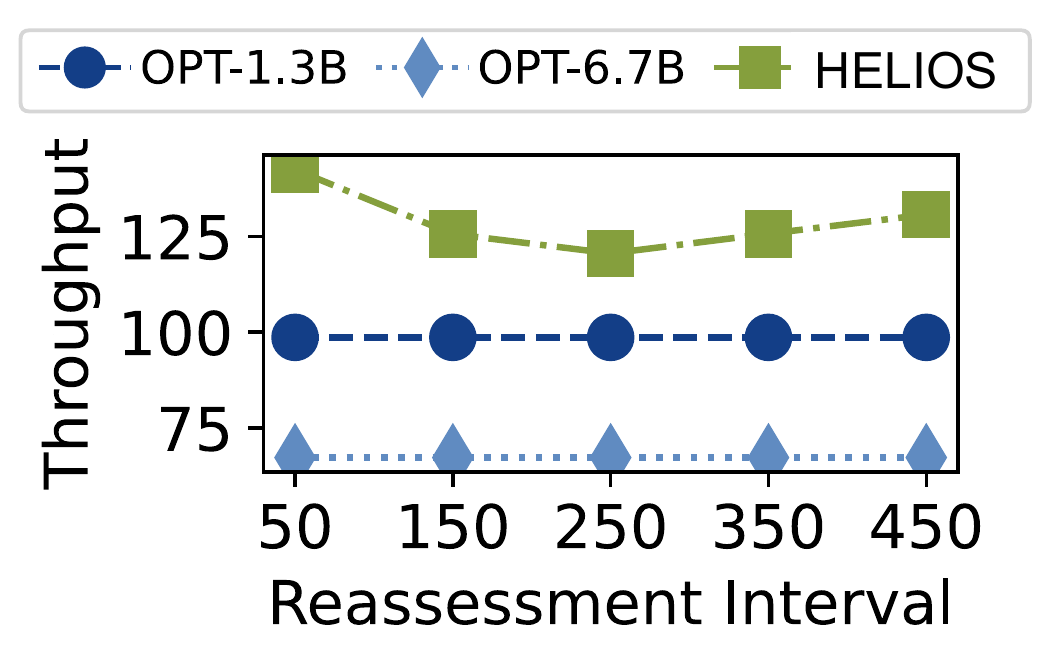}
   \vspace{-0.2in}
   \caption{Impact of increasing $RI$ on throughput.}\vspace{-0.1in}
   \label{fig:variableCT}
\end{figure}

The Re-assessment Interval ($RI$) hyper-parameter impacts the performance of \ours-- too low values of $RI$ triggers frequent re-evaluation of the early-exit profiles and incur overheads, whereas large values imply that \ours may be potentially serving requests with obsolete exit profiles. Figure~\ref{fig:variableCT} shows the throughput for increasing values of $RI$. 
Existing EE-LLM frameworks do not have any notion of $RI$ and thus, their throughput remains unimpacted. 
For \ours, the throughput is highest for $RI$=50 and reduces up to 250. Also, higher values of $RI$ yields higher throughput but there is also a high likelihood that this impacts accuracy, because the nature of the incoming prompts may change between two re-assessment phases. These may go undetected if \ours never exceeds the Confidence Breach Counter during this timeframe. Our default implementation uses an $RI$ of 150 because \ours also involves additional model switching due to the Confidence Breach Counter exceeding its maximum tolerable limit in between re-assessment intervals. Thus, by picking an $RI$ slightly above 50 allows us to minimize the overheads of re-assessment without largely impacting throughput and accuracy. 

\ignore{
Table~\ref{tab:energyvsTH} shows the energy per prompt with varying $TH$ values. As expected, the energy per prompt increases with increasing $TH$. However, as \ours uses more early exits from both models combined, its energy-efficiency is considerably higher.

    \vspace{-0.1in}
    \begin{table}[!htb]
        \centering
        \begin{center}
        \caption{Energy per prompt (in mWh) with increasing $TH$} \vspace{-0.1in}
        \label{tab:energyvsTH}
        
        \renewcommand{\arraystretch}{1.2}
        \begin{tabular}{|c|c|c|c|c|c|}
            \hline
            Model & 0.5 & 0.6 & 0.7 & 0.8 & 0.9 \\
            \hline
            \hline
            OPT-1.3B & 124 & 136 & 146 & 159 & 182\\
            \hline
            OPT-6.7B & 253 & 266 & 277 & 296 & 321\\
            \hline
           \ours & 113 & 122 & 124 & 123 & 117\\
            \hline
        \end{tabular}
        \vspace{-0.15in}
        \end{center}
    \end{table}
}

\subsection{Generalization to Other SLOs}
The results presented in this section so far considers the user default SLO is to maximize throughput. However, \ours is generalizable and can effectively accommodate a wide-range of SLO objectives. To evaluate this, we also consider other SLOs, such as response time, accuracy, and energy-efficiency. We provide a comprehensive analysis of our results in the Appendix~\ref{sec:otherslos}.

\section{Related Work}
Prior software and hardware works explore trade-offs across LLM performance metrics, we compare and contrast below:

\vspace{0.05in}
\noindent \textbf{Comparison with Speculative Decoding:} {Speculative Decoding}~\cite{leviathan2023fast, kim2023speculative, stern2018blockwise} is an inference-time technique which orchestrates two models, a smaller draft model, and a larger target model. The draft model generates output tokens one-by-one, while the target model periodically verifies and corrects them in parallel. However, unlike speculative decoding, early-exits significantly reduce energy-consumption by skipping further layers of the same model, and eliminating the need for a computationally intensive verification phase. Our experiments indicate that a speculative system consisting of OPT-125M, and OPT-6.7B consumes 1.49$\times$ more energy compared to OPT-6.7B with two early exits on the CNN Daily Mail~\cite{cnn-daily-mail} dataset. LayerSkip~\cite{elhoushi2024layerskip} incorporates early-exits into their speculative decoding framework for further benefits. 
Furthermore, the selected draft layer in Layerskip remains static throughout execution, regardless of variations in the nature of inputs within the request stream.~\ours{} adaptively switches between EE-LLM models or loads more layers to collectively maximize early-exits. Single exit-layer based speculative decoding can still be adopted by~\ours{} if the served model being served contains atleast one intermediate exit prior to reaching the depth of the greedily loaded EE-LLM.

\vspace{0.05in}
\noindent \textbf{Hardware-Software Co-Design:} BERT Loses Patience \cite{bert-patience} proposes a technique for fast and robust inference by introducing a patience-based early exit. In this approach, auxiliary classifiers are attached to intermediate transformer layers, and the forward pass dynamically terminates if the predicted class remains the same across a predefined number of consecutive layers. This approach leverages several intermediate layers in order to terminate the forward pass for a low-confidence token. In contrast,~\ours{} uses early-exit distributions in real-time to greedily only loads the most likely to be used layers in GPU memory. This allows~\ours{} to not only improve the token-generation latency, but also support larger batch sizes due to the memory savings yielded by its greedy approach. Prior work,
Edge-BERT~\cite{edge-bert} proposes exploiting GPU power-management (DVFS) to adaptively manage the GPU clock frequency based on predicted early exits. This allows Edge-BERT to save substantial power when tokens exit after traversing very few transformer layers.~\ours is a generalized framework capable of accommodating any user-defined SLO, rather than being restricted to energy-efficiency, which is the primary focus of Edge-BERT.

\vspace{0.05in}
\noindent \textbf{Model Serving Optimizations:} In addition, there have been several works on LLM inference serving at the cloud. 
INFaaS~\cite{romero2021infaas} selects a model that meets SLOs of the task and performs inference with it, and Clipper~\cite{crankshaw2017clipper} combines predictions from multiple models hosted concurrently. \ours instead dynamically selects a model to get predictions from a single model at a time, while ensuring the overall perplexity remains similar. Techniques like pipeline parallelism~\cite{narayanan2019pipedream}, and model parallelism, as used in AlpaServe~\cite{alpaserve} are complementary to \ours, and could be combined to scale our design to accommodate larger models on the GPUs. Similarly Splitwise~\cite{patel2024splitwise}, a technique which splits the prefill and the token generation phase across multiple GPUs complements the design of \ours.


\noindent \textbf{Dynamic Neural Networks:} Multiple works have developed neural networks that have different computational graphs, based on particular deployment and prompt scenarios~\cite{dynamicnn,convnet-aig}. However unlike \ours, these frameworks only consider a single model and the configurations within it.

\section{Conclusion}
Early-Exit LLMs (EE-LLMs) are promising variants of LLMs that enable high throughput inference by allowing tokens to exit at specific intermediate layers during the forward pass if their probability meets a predefined confidence threshold. This makes them attractive as they improve throughput without compromising accuracy. Existing EE-LLM frameworks rely on a single model and thus, their token generation latencies are primarily limited by the tokens that do not exit early. To accommodate the worst case exit-depth, current EE-LLM serving frameworks load the weights of all model layers, even though the memory corresponding to the later layers remain unused when tokens exit early. Limited latency savings and poor memory management severely limits us from attaining large throughput benefits in these frameworks.

In this paper, we propose~\ours{}, an adaptive EE-LLM serving framework that yields reduced token generation latencies by maximizing the total number of early-exit tokens; and improves memory efficiency, enabling us to scale batch sizes. \ours orchestrates multiple models and dynamically switches between them to collectively maximize early-exits for a given set of input prompts. Furthermore, it exploits the insight that low-confidence tokens that do not take early exits often remain unchanged even after additional layer traversal. Therefor, \ours{} leverages greedily loads only the weights of the most likely to be used layers onto the GPU memories, yielding memory savings, which are then repurposed to support larger batch sizes. Our studies show that~\ours{} achieves $1.48\times$ throughput and up to $15.14\times$ higher batch size compared to existing frameworks while meeting SLOs with negligible impact on accuracy.

\section*{Acknowledgments}
The authors acknowledge the Texas Advanced Computing Center (TACC) and the Center for Generative AI at the University of Texas at Austin for providing computational resources that helped develop the research results reported in this paper. This research was supported in part by NSF Grants \#2326894 and \#2425655, and the NVIDIA Applied Research Accelerator Program Grant. Poulami Das acknowledges the generous support through the AMD endowment at the University of Texas at Austin.
\bibliography{references}
\bibliographystyle{mlsys2025}

\clearpage
\appendix

\section{Using Perplexity for Accuracy}
\label{app:perplexity}

~\ours{} performs inference-time accuracy estimation to decide when to selectively load more layers or switch to a different model. Accuracy estimation methods can be grouped into three-main categories:
\begin{itemize}
    \item \textbf{Reference-Based Metrics:} This category of metrics rely on pre-defined ground truth labels to evaluate the quality of generated text. They measure the degree of overlap between the generated and the ground-truth reference using lexical or semantic similarity measures such as ROUGE, BLEU, and METEOR.
    
    \item \textbf{Reference-Free Metrics:} These metrics evaluate the quality of generated text without relying on reference outputs. Instead, they asses aspects such as fluency and coherence directly from the model output itself, using measures like Perplexity, BLANC, and Supert.

    \item \textbf{LLM-Based Metrics:} These metrics leverage LLMs for evaluating the quality of generated text. LLM-based evaluators are typically prompt-driven and can operate in both reference-free and reference-based scenarios. Common frameworks include Reason-then-Score (RTS), Head-to-Head (H2H), and G-Eval.
\end{itemize}

As~\ours{} is an inference-time technique and ground-truths for incoming requests are unavailable to the server, it is limited to employing reference-free methods for accuracy estimation. While LLM based-metrics provide with detailed assessments of output quality, their usage is impractical because they require loading a separate, much larger LLM in memory than the one being evaluated.

We use \emph{Perplexity} to estimate accuracy in real-time. Unlike metrics such as BLANC which measure how the output text aids the language model to reconstruct the source document, Perplexity captures a models own confidence over its output distribution. BLANC involves repeated masking and re-encoding operations, making it unsuitable for real-time inference. In contrast, Perplexity is computed directly from token probabilities, providing an intrinsic, reference-free measure of output quality that can be efficiently computed during inference. Furthermore, our evaluations ensure that perplexity comparison between models is fair by restricting the model repository to only those models which have the same tokenizer and vocabulary, thereby maintaining consistency in tokenization and likelihood normalization.

\section{Benchmarks Used for Evaluation}
\label{app:benchmarks}

We evaluate~\ours{} across a wide-range of LLM tasks including conversation, summarization, mathematical reasoning, code-generation, and sentence completion. The following section provides detailed descriptions of the benchmarks employed in our evaluations:

\textbf{Conversation:} The ShareGPT dataset~\cite{sharegpt} is a collection of publicly available user-shared conversation logs between humans and a LLM. These logs have been used by various open-source initiatives for training and evaluating LLMs. In particular, we adopt the same version utilized in training the Vicuna~\cite{vicuna2023} family of models.

\textbf{Summarization:} The CNN/Dailymail dataset~\cite{cnn-daily-mail} is a large-scale summarization benchmark. It comprises news articles sourced from the CNN and the Dailymail website. Each article is accompanied with highlights that capture the key-points of the article.

\textbf{Code-Generation:} The CodeXGLUE dataset~\cite{codexglue} is a comprehensive benchmark to measure both code understanding and generation. It encompasses a variety of tasks including code completion, summarization, translation, and bug detection, spanning multiple programming languages such as Python, Java, Javascript, and C\#. 

\textbf{Mathematical Understanding:} To asses arithmetic and reasoning capabilities, we employ the GSM8K dataset~\cite{gsm8k}. It consists of 8.5K grade school level math word problems. Each problem requires multi-step reasoning to arrive at the numerical answer, testing the LLMs ability to deduce logic and perform basic mathematical computation.

\textbf{Sentence Completion:} The HellaSwag~\cite{hellaswag} dataset is designed to test the sentence completion ability of a LLM. Each example provides a short-narrative followed by many continuations, only one of which forms a coherent and context-appropriate completion.

\begin{figure}[!htp]
   \centering
   \includegraphics[width=\linewidth]{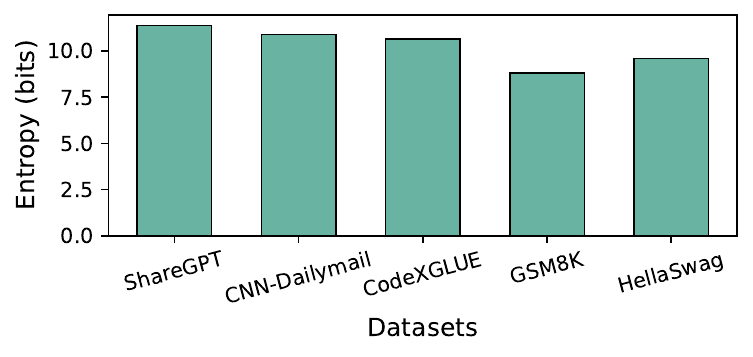}
   \caption{Comparison of dataset entropy. Higher entropy indicates lower predictability and greater diversity in lexical structure.}
   \label{fig:dataset_entropy}
\end{figure}

Figure~\ref{fig:dataset_entropy} compares the token-level entropy across these datasets. Higher entropy indicates lower predictability and greater lexical diversity, while lower entropy corresponds to higher predictability and a more repetitive structure. These datasets exhibit consistently high entropy, suggesting that accurately predicting output tokens is non-trivial. Overall, the usage of these datasets ensures that our evaluation maintains a consistent level of predictive challenge.

\begin{figure*}[!t]
  \centering
  \includegraphics[width=\linewidth]{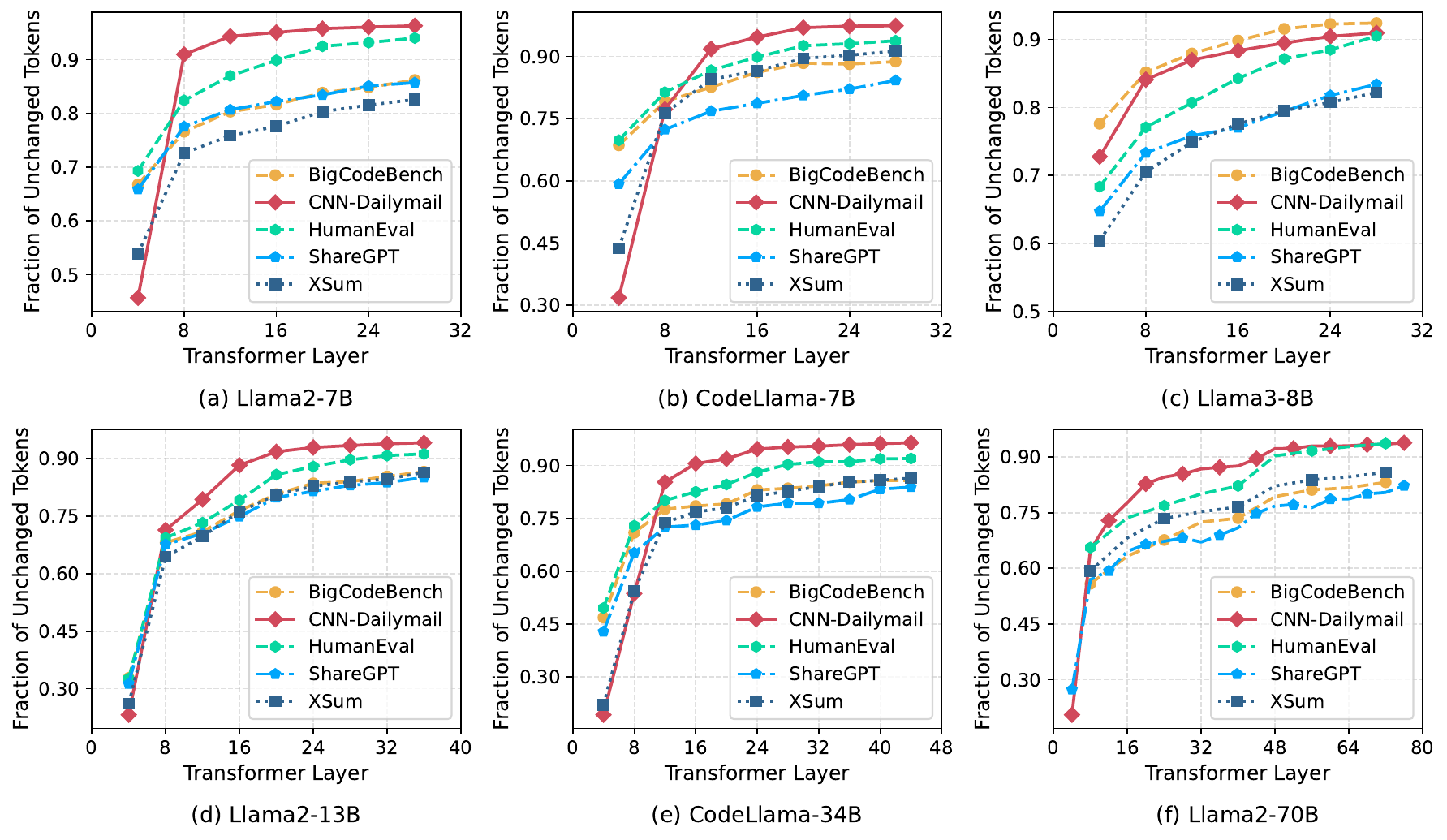}
   \caption{Fraction of unchanged tokens produced by an exit layer that remain unchaged even after additional model traversal. We observe that the probability that a predicted token remains unchanged reaches up to 90\% for the Llama2-7B model as early as layer 8 i.e., one fourth the model depth. Furthermore, this trend remains consistent across models of various sizes from the Llama family.}
   \label{fig:llama_exit_prof}
\end{figure*}

\section{Low-Confidence Tokens Remain Unchanged}
\label{app:early_good}

Transformer-based LLMs are composed of several decoder layers. Each layer processes the output of the previous layer to progressively refine the output token. EE-LLMs introduce mechanisms that enable output tokens to exit at an intermediate layer if a predefined confidence criteria is satisfied. However, we observe that typically, low-confidence tokens which fails to take early-exit, remain unchanged even after additional model traversal. For example, Figure~\ref{fig:llama_exit_prof} shows the fraction of tokens that are produced by an exit layer which remain unchanged even after traversing the full-model. We observe that in the case of the Llama2-7B~\cite{llama-2} model, the eighth layer (one quarter of the model depth) accurately produces upto 90\% of the tokens for the the CNN-Dailymail~\cite{cnn-daily-mail} dataset. Furthermore, this trend remains consistent across models of various sizes from the Llama family~\cite{llama-2,llama-3,codellama}, indicating that early-exit layers possess strong predictive capabilities irrespective of model scale.~\ours{} leverages this insight to greedily load only the most likely to be used layers, which yields memory savings. These memory savings are then repurposed to support support larger batch sizes.

\section{Evaluated Models}
\label{app:models}

We evaluate~\ours{} using both pre-trained and post-training modified EE-LLMs. Pre-trained models are trained to allow intermediate exits directly during the models original training process, while post-training modified models refer to standard off-the-shelf LLMs that are later augmented with early-exits through fine-tuning. The remainder of this section provides specifics for each category:

\textbf{Pre-trained models:} Layerskip~\cite{elhoushi2024layerskip} has publicly released several pre-trained early-exit models on HuggingFace~\cite{huggingface}, spanning the Llama2~\cite{llama-2}, Llama3~\cite{llama-3} and CodeLlama~\cite{codellama} families. These models are trained to allow tokens to exit after any arbitrary layer, offering greater flexibility for~\ours{} in greedily loading only the most likely to be used model layers. Compared to their standard counterparts, Layerskip variants achieve comparable performance while supporting early-exits.

\textbf{Post-training modified models:} We extend two base models from the OPT family by inserting early-exit mechanisms (auxiliary heads) at selected intermediate layer depths. These models are then fine-tuned with the backbone parameters kept frozen, while only the parameters of the auxiliary heads being updated in every iteration. Prior work~\cite{xin2021berxit,ee-tune} has observed that such selective fine-tuning preserves the output generation ability of the backbone model while enabling efficient early-exits through auxiliary heads. The total loss is computed by summing the loss at each exit-layer with a weight of 1.0. For fine-tuning, we utilize the RedPajama~\cite{redpajama} and the Pile~\cite{the-pile} datasets over a total of 50K iterations. Table~\ref{tab:models} summarizes the models used in our evaluations:

\begin{table}[ht]
\centering
\caption{Overview of models used for evaluating~\ours{}. Pre-trained models permit tokens to exit at any arbitrary layer, while post-training modified models restrict exits to specific layers\label{tab:models}.\\} 

\begin{tabular}{lccc}
\toprule
Family & Parameters & Depth & Early-Exits \\
\midrule

\rowcolor{mycolor}
\multicolumn{4}{l}{\texttt{Pre-trained}} \\
\quad \multirow{3}{*}{Llama2}
& 7B & 32 & \multirow{5}{*}{Anywhere} \\
& 13B & 40 & \\
& 70B & 80 & \\
\quad Llama3 & 8B & 32 & \\
\quad CodeLlama & 34B & 48 & \\
\midrule

\rowcolor{mycolor}
\multicolumn{4}{l}{\texttt{Post-training modified}} \\
\quad \multirow{2}{*}{OPT}
& 1.3B & 24 & 6, 12, 24 \\
& 6.7B & 32 & 9, 17, 32 \\

\bottomrule
\end{tabular}
\end{table}

\begin{figure*}[tp]
  \centering
  \includegraphics[width=\linewidth]{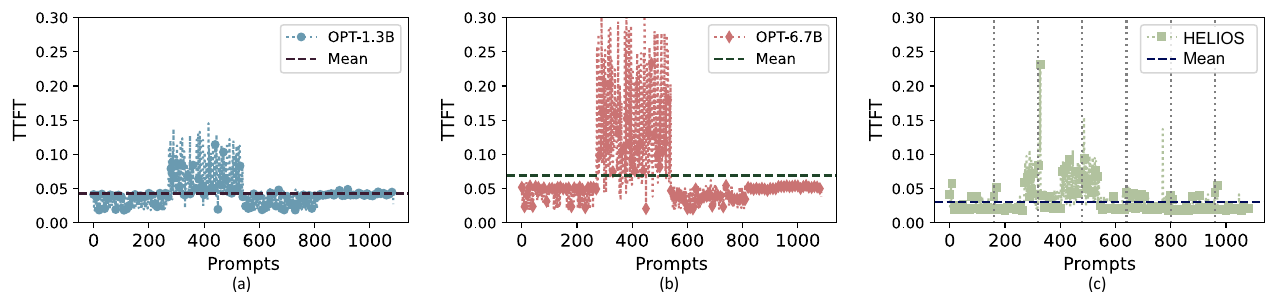}
   \caption{Comparison of\textit{ Time To First Token (TTFT)} when using (a) only OPT-1.3B, (b) only OPT-6.7B, and (c) \ours (\textit{lower is better}). In (c), the vertical lines denote timestamps when a candidate re-assessment is initiated. }  
   \label{fig:ttft}
\end{figure*}

\section{Additional Results}
\label{sec:otherslos}
~\ours{} is a unified and flexible framework capable of optimizing for any user-defined SLO. In this section, we show the effectiveness of~\ours{} for three SLOs: response-time, accuracy, and energy-efficiency.

\subsection{Response Time}\label{app:response-slo}
When the user's objective is to minimize response time, we evaluate the \textit{Time Taken to First Token or TTFT (lower is better)}, shown in Figure~\ref{fig:ttft}. TTFT is limited by the time taken to process all the input tokens in the request. During this time, no output tokens are produced. As expected, larger models, like OPT-6.7B, incur a higher TTFT (69ms) compared to the smaller OPT-1.3B model (43ms). In contrast, \ours achieves $1.39\times$ and $2.23\times$ reduction in TTFT compared to OPT-1.3B and OPT-6.7B respectively. This is expected as \ours greedily loads only the most likely to be used layers. With \ours, each input token in the request traverses fewer layers, significantly reducing the time spent in processing input tokens. Also, \ours maximizes the total number of requests served using early exits from both models combined. This is particularly evident in Figure~\ref{fig:ttft} for requests 272 to 542 which corresponds to the CNN-Dailymail dataset comprising long input tokens.~\ours outperforms the OPT-6.7B model by up to $30\times$ for some of these requests due to reduced layer traversal.

\begin{figure*}[tp]
  \centering
  \includegraphics[width=\linewidth]{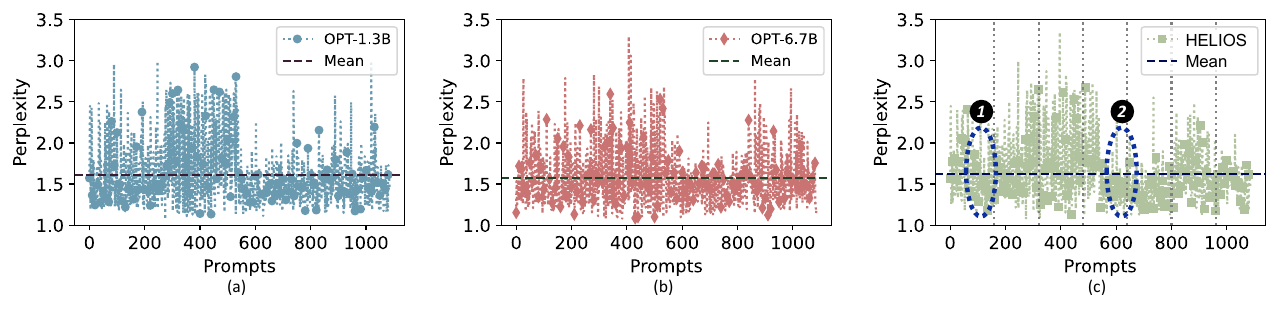}
   \caption{Comparison of \textit{Perplexity (Accuracy)} when using (a) only OPT-1.3B, (b) only OPT-6.7B, and (c) \ours (\textit{lower is better}). Vertical lines represent timesteps when a candidate re-assessment is initiated.}  
   \label{fig:accuracy} \vspace{-0.15in}
\end{figure*}

\subsection{SLO: Accuracy Optimization}\label{app:acc-slo}
Next, we study the scenario when the user's goal is to maximize the accuracy. Figure~\ref{fig:accuracy} shows the \textit{perplexity (lower is better)} for three model selection cases. The perplexity of the larger model, or the OPT-6.7B model, is lower ($0.97\times$) than the smaller OPT-1.3B model. This is expected because larger models typically have many parameters enabling them to better capture the subtle relationships between tokens and produce more nuanced outputs. Specifically, for long context benchmarks, where a significantly large number of output tokens are generated, larger and complex models are known~\cite{achiam2023gpt,kaplan2020scaling} to outperform smaller models. Hence, for the set of prompts corresponding to CNN Daily Mail~\cite{cnn-daily-mail}, which comprises relatively long context inputs, \ours switches to using OPT-6.7B in real-time, as illustrated by \bcircled{1} in Figure~\ref{fig:accuracy}(c), to meet the target SLO of the user. This highlights the ability of \ours to adapt to various SLO requirements.
On the other hand, \ours reverts back to OPT-1.3B later, as shown by \bcircled{2} in Figure~\ref{fig:accuracy}(c),
because \ours evaluates that it offers accuracy comparable to OPT-6.7B in a more energy-efficient manner (with fewer layers and reduced computational footprint).

\subsection{SLO: Energy-efficiency Optimization}\label{app:energy-slo}
We briefly discuss the scenario when a user wants to maximize the energy-efficiency or \textit{minimize the energy per prompt}. 
Using only OPT-6.7B consumes 1.01 Wh of energy per prompt which is expected given it is a larger model compared to OPT-1.3B that consumes 0.50 Wh per prompt. In contrast, \ours consumes 0.45 Wh of energy per prompt, which translates to 10\% energy savings, for comparable perplexity. In \ours,
58.3\% of the prompts are serviced using partially loaded models , which yields the observed energy savings. Note that savings scale with the total number of prompts processed. In practice, production servers in datacenters process tens of millions of prompts daily~\cite{wang2023survey}, emphasizing the impact of \ours. 
We also observe that the energy overheads associated with switching is minimal, comprising only $0.05\times$ of the overall energy savings (10\%) achieved.

\end{document}